\documentclass{article}
\usepackage{spconf,amsmath,graphicx,hyperref}
\usepackage{multirow}
\usepackage{subcaption}
\usepackage{comment}
\usepackage{enumitem}
\usepackage{hyperref}

\title{A Dataset of Robot-Patient and Doctor-Patient Medical Dialogues for Spoken Language Processing Tasks}
\name{Author(s) Name(s)}
\address{Author Affiliation(s)}
\twoauthors
  {Heriberto Cuay\'ahuitl} 
	{School of Engineering and Physical Sciences\\
	University of Lincoln, UK}
  {Grace Jang} 
	{Lincoln Medical School\\
	Universities of Lincoln and Nothingham, UK}
\begin{document}
%
\maketitle
\begin{abstract}
Large Language Models (LLMs) have brought huge improvements to Artificial Intelligence (AI), which can be applied to general-purpose tasks. However, their application to textual or spoken medical consultations is still an open research problem. This paper proposes \href{https://huggingface.co/datasets/hcuayahu/MeDial-Speech}{MeDial-Speech}, a novel speech dataset for training and evaluating Med-AIs that can carry out consultations with patients. It was collected in realistic environments from robot-patient and doctor-patient dialogues, contains 111+ hours of speech data (without data augmentation), and covers four health conditions: Lewy body dementia, heart failure, shoulder pain, and angina. In addition, we propose a dialogue benchmark via sentence selection (with 20 options) to evaluate three state-of-the-art LLMs: GPT-5 mini, DeepSeek-V3, and Claude Sonnet 4. Experimental results reveal that Claude Sonnet 4 is the best in sentence selection, with 71.1\% accuracy using manual transcriptions and 74.7\% using automatic transcriptions, and that all LLMs are highly overconfident in their probabilistic predictions, regardless of selecting correct or incorrect sentences in medical dialogues. This dataset is free of charge for non-commercial purposes.
\end{abstract}
\section{Motivation}
\label{sec:intro}
\begin{figure}[t]
\centering
\includegraphics[width=1.0\columnwidth]{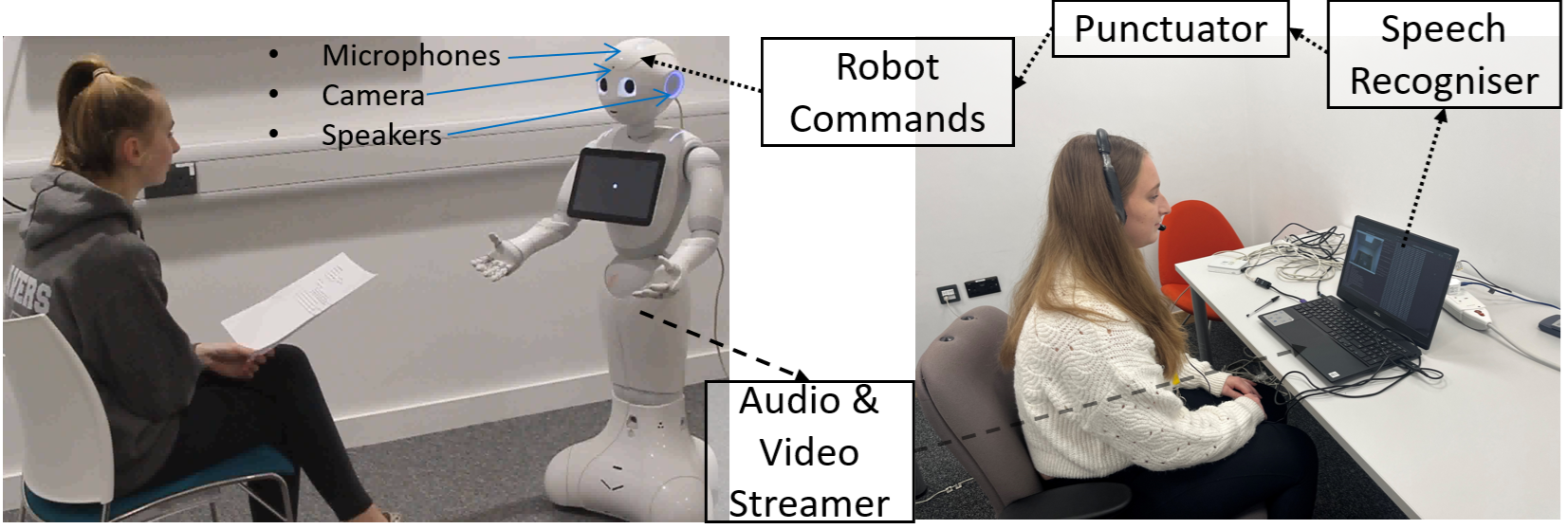}
\caption{An actor patient interacting with a teleoperated robot.}
\label{robot-patient-consultation}
\end{figure}

Effective communication between doctors and patients is an essential aspect of the UK's NHS core value, ‘commitment to quality of care’\cite{NHS2025}. 
Exposure to medical consultations is crucial for trainees---whether human or machine---in the medical field to enhance their competence in communication, clinical reasoning, and professionalism.
This paper introduces a new dataset of doctor-patient and robot-patient medical dialogues, with the intention of aiding both medical AIs and students in clinical skills training outside of clinical placement. In addition, this dataset aims to contribute to the development of automated clinicians, as in Fig.~\ref{robot-patient-consultation}, with the ability to manage non-urgent conditions. These types of efforts have the potential to alleviate the workload of healthcare professionals since human clinicians can direct their attention towards patients with more critical needs or other attention-demanding tasks.

\section{Previous Works}
\label{sec:lit_review}

Speech-based medical datasets publicly available are scarce. Some exceptions include \cite{SnaithEtAl-LREC2021}, who used four participants for a total of 9 consultations with audio and video recorded, \cite{FareezEtAl-Nature2022}, who collected and annotated 272 human-human consultations, \cite{Papadopoulos-Korfiatis22}, who reported a dataset of 57 primary care consultations, \cite{barnes2017one}, who collected and annotated 327 human-human consultations, and \cite{definedai_med_dialogues}, who reported 2,000 hours of speech. Whilst only the latter two involve real patients, they are not publicly available and require purchase. 

Automatic Speech Recognition (ASR) efforts have been made in this realm, but the speech data is either closed source \cite{chiu2018speech} or only from human-human conversations with data/synthetic augmentation \cite{adedeji2024sound,le2024multimed}. Their performance ranges between 12\% and 30\% in word error rate, and these ASR models have not been evaluated using interactive systems and human participants. To our knowledge, there is no publicly accessible, free-of-charge dataset of spoken medical consultations that includes both human–machine and human–human interactions (and with multiple levels of annotations), both of which are arguably valuable for developing an AI doctor. 

Medical LLMs, on the other hand, are trained from large text-based datasets such as \cite{johnson2016mimic,johnson2022mimic,han-etal-2024-medinst}. Whilst they have shown remarkable abilities in classification and discriminative tasks \cite{zhou2024surveyllms,ZhengGCQLY25}, they are not specialised for interactive medical consultations. Nonetheless, medical LLMs and speech-based medical data (as proposed in this work) will play a crucial role in the development of future conversational AIs, including robots for healthcare delivery \cite{ValizadehEtAl-ACL2022,SawadEtAl-Sensors2022,KyrariniEtAl-Technologies2021,DinglerEtAl-IMIA2021}.

\section{The Data Collection System}
\label{sec:system}
The Pepper robot \cite{pandeyEtAl2018} was equipped with a teleoperated system using in-person and remote embodied telepresence. In this setup, the teleoperator was embodied in the body of the robot by hearing and seeing what the robot perceives and by saying what the robot conveys to the person in front of it. This Wizard-of-Oz (WOZ) setting is illustrated in Figure~\ref{robot-patient-consultation}, where the human doctor (teleoperator) used a separate room from the robot and the patient. 
We avoided the use of GUIs as in \cite{rietzEtAl2021,bonialEtAl2018} to prevent predefined responses and to encourage personalised, human-like communication.

Instead of using the teleoperator's actual speech, we used continuous speech recognition via \href{https://alphacephei.com/vosk/}{Vosk} (model: vosk-model-en-us-0.22, 2.6GB), together with automatic punctuation of \href{https://github.com/notAI-tech/fastPunct}{FastPunct} and speech synthesis via \href{https://www.acapela-group.com/}{Acapela}. This means that as the teleoperator uttered a sentence, automatic speech recognition generated words actively and indefinitely. Whenever a two-second pause was detected, the recognised and generated sentence was automatically punctuated. The latter was useful for producing more intelligible speech than without punctuation. 
After punctuation, a motor command was generated to execute robot speech concurrently with body movements (gestures: chosen randomly from a set of 18 pre-recorded motions of robot arms and head related to greeting, explain, yes, no, and please). 

Other non-verbal behaviours included face tracking (to follow patient gaze) and head nodding (for acknowledgements) during human speech. The medical doctors---all medical students---were able to hear and see at all times during a dialogue via continuous audio and video streaming on the laptop in front of them. Additional technical features in this robot system are: (1) audio recorded with sampling rate at 16 kHz, 16-bit, and 1 channel; (2) audio produced at 44 kHz with 2 channels, TTS system with speed=100, pitch shift=1.5, and volume=60\%; (3) video at 10 images per second, 640x480 pixels; (4) robot and laptop connected via Ethernet.

The dynamics of our conversational robot system are briefly described as follows. The teleoperator's speech was automatically transcribed via ASR and punctuation, then used to produce robot speech via speech synthesis. Robot speech is what the patients were able to hear, i.e., they never heard the actual speech of the human doctor. The patient's speech and video were continuously captured by the robot without interruption during a consultation. These two streams of data were perceived by the human doctor in near real time through headphones and the laptop screen. The latter displayed the patient's video and the transcriptions of the doctor’s speech. This procedure is illustrated in Figure~\ref{robot-patient-consultation}, which began by specifying the participant ID and then pressing Enter to indicate the start of a consultation. A consultation terminated once the teleoperator said the phrase “let’s move on.” Finally, 
human-human consultations were recorded using \href{https://www.microsoft.com/en-gb/microsoft-teams/group-chat-software}{MS Teams}. 

\section{The Speech Dataset}
\label{sec:dataset}

\subsection{Data Collection}
We collected data from 325 recruited and unpaid participants, mostly university students and from different schools (but mainly the Schools of Medicine and Computer Science at \href{https://www.lincoln.ac.uk/}{UoL}) with age categories 18-24 (87.1\%), 25-34 (8.9\%), and 35+ (4.0\%). All participants were fluent speakers of English, including native \& non-native speakers of both genders. Each participant read and signed a consent form and reviewed a task description and a simulated patient profile prior to the consultation. Patient profiles included the following health conditions: Lewy-body dementia, heart failure, shoulder pain, and angina. Since our participants were simulated patients instead of real ones, they were allowed to keep a printed sheet describing their patient profile during the consultation. Afterwards, each participant completed a questionnaire of 10 questions with a 5-point Likert scale and three free-text questions for open-ended comments. Whilst the results of the questionnaires will be reported in a separate paper comparing different versions of in-person and remote conversants, statistics of our proposed dataset are listed in Table~\ref{stats}. 

\begin{table}[!t]
\small
\centering
\begin{tabular}{ |l|r| } 
 \hline
Total Duration (in hours) & 111.4\\
Total File Size (in Gigabytes) & 12.6\\
Total Number of Dialogues & 581\\
Total Number of Dialogue Turns & 11197\\
Total Number of Words & 264451\\
Avg. Num. of Turns per Dialogue & 22.48\\
Avg. Num. of Words Per Dialogue & 531.03\\
 Total Number of Unique Words (with contractions) & 6100\\
 Total Number of Unique Doctor Sentences & 7987\\
 Total Number of Unique Patient Sentences & 8206\\
 \hline
\end{tabular}
\vspace{-2.0mm}
\caption{\label{stats}Indicative statistics of the proposed dataset.}
\vspace{-3.3mm}
\end{table}

\begin{table*}[h!]
\small
\centering
\begin{tabular}{ |l|c|c|c||c|c|c| } 
 \hline
\multirow{2}{*}{\centering Metric} &  \multicolumn{3}{c||}{Without Noise (manual transcriptions)} & \multicolumn{3}{c|}{With Noise (ASR transcriptions)}\\
\cline{2-7}
                                          & GPT-5 mini & DeepSeek V3 & Claude Sonnet 4 & GPT-5 mini & DeepSeek V3 & Claude Sonnet 4 \\ 
 \hline
 \hline
Balanced Accuracy ↑ & 0.4919 & 0.6271 & {\bf 0.7119} & 0.5054 & 0.5598 & {\bf 0.7473} \\
F1 Score ↑ & 0.6591 & 0.7708 & {\bf 0.8317}  & 0.6715 & 0.7178 & {\bf 0.8554} \\
Area Under Curve (AUC) ↑ & 0.5119 & 0.6291 & {\bf 0.6897}  & {\bf 0.6852} & 0.5479 & 0.6709 \\
\hline
Brier Score ↓ & 0.2754 & 0.2421 & {\bf 0.1888}  & 0.2450 & 0.3069 & {\bf 0.1917} \\
KL Divergence (total) ↓ & 267.6951 & 245.3904 & {\bf 198.4698}  & 284.3294 & 359.5171 & {\bf 208.2200} \\
Binary Cross Entropy ↓ & 0.7562 & 0.6932 & {\bf 0.5606}  & 0.7726 & 0.9769 & {\bf 0.5658} \\
Expected Calibration Loss ↓ & 0.1119 & 0.1321 & {\bf 0.0541}  & 0.1446 & 0.2526 & {\bf 0.1151} \\
 \hline
\end{tabular}
\vspace{-2.0mm}
\caption{\label{llm_metrics} LLM performance in sentence selection using classification and probability metrics: ↑=higher is better, ↓=lower is better. Best results in bold. ASR transcriptions were applied only to patient utterances, to simulate real-world conditions.}
\vspace{-3.3mm}
\end{table*}

\subsection{Data Annotations}
The data has been annotated with respect to speech transcriptions and speaker roles (robot, doctor, patient, or none). 
Here is an example dialogue: \url{https://youtu.be/9z8KDRh_mn4}, which uses our own tool to replay consultations in three versions--in terms of conversants involved: robot-patient, doctor-patient, or doctor-robot-patient\footnote{The medical consultation player was implemented as a local web server using .NET SDK 6.0 and Node.js v20 or later.}. Each consultation in this dataset is stored in a separate folder containing the following: (a) raw audio data stored in two files *.wav due to using two separate rooms, one file for the teleoperator and another for the robot \& patient speech; (b) speech transcriptions stored in a text file denoted as *-3speakers.txt with the speech transcriptions for the three speakers (in the case of robot-human consultations), two speakers otherwise (in the case of human-human consultations); and (c) \href{https://www.audacityteam.org/}{Audacity} transcriptions stored in two audacity files *.aup3 include the annotations of speaker roles (one for each audio file), and text files generated from those audacity files containing the start and end times for each segment of speech. 
The raw audio files were split and stored in subfolders (one per speaker) containing audio files for each dialogue turn in a consultation. The annotations have been revised to verify that the number of dialogue turns in the textual transcriptions matches the speech segments, allowing consultations to be replayed in the standalone application shown in the example dialogue above.

\section{Data Uses: Benchmarks and Tools}
\label{sec:benchmarks}
The \href{https://huggingface.co/datasets/hcuayahu/MeDial-Speech}{MeDial-Speech} dataset can be used for benchmarking the following Spoken Language Processing tasks. 
We focus on making data available with reference transcriptions and labels, and leave data splits of training, validation, \& test open. 

1. {\bf Voice Activity Detection (VAD)}. This task can be performed, for example, by randomly sampling segments of speech, where the annotations (speech, non-speech) can be generated from the Audacity transcriptions. If a speech segment falls mostly within any speaker (doctor, robot, patient), then it should be labelled as {\it speech}; {\it non-speech} otherwise.

2. {\bf Automatic Speech Recognition (ASR)}. This task can be addressed using an ASR system of interest in two ways: (i) using the audio splits from the previous section, one per utterance; and (ii) using the raw audio files, one per consultation. The former compares a newly generated speech transcription (predicted text) against the reference transcription for that dialogue turn. 
The latter allows a similar comparison but over the entire consultation rather than a single dialogue turn.

3. {\bf Dialogue Generation}. This task can be studied in at least two ways: (i) generating the next response given a context dialogue snippet ({\it generation task}); and (ii) predicting the next response from a set of options given a context dialogue ({\it classification task}), similar to multi-choice question answering or sentence selection. Whilst the generation task produces an open-ended response, the classification task only selects the most relevant response from the available options.

4. {\bf Educational Tools}. Medical students and academics can benefit from this dataset by analysing the consultation dialogues and identifying positives and negatives to learn from.

\begin{figure*}[h!]
    \centering
    \begin{subfigure}[b]{0.45\textwidth}
        \flushright
		\includegraphics[width=0.99\columnwidth]{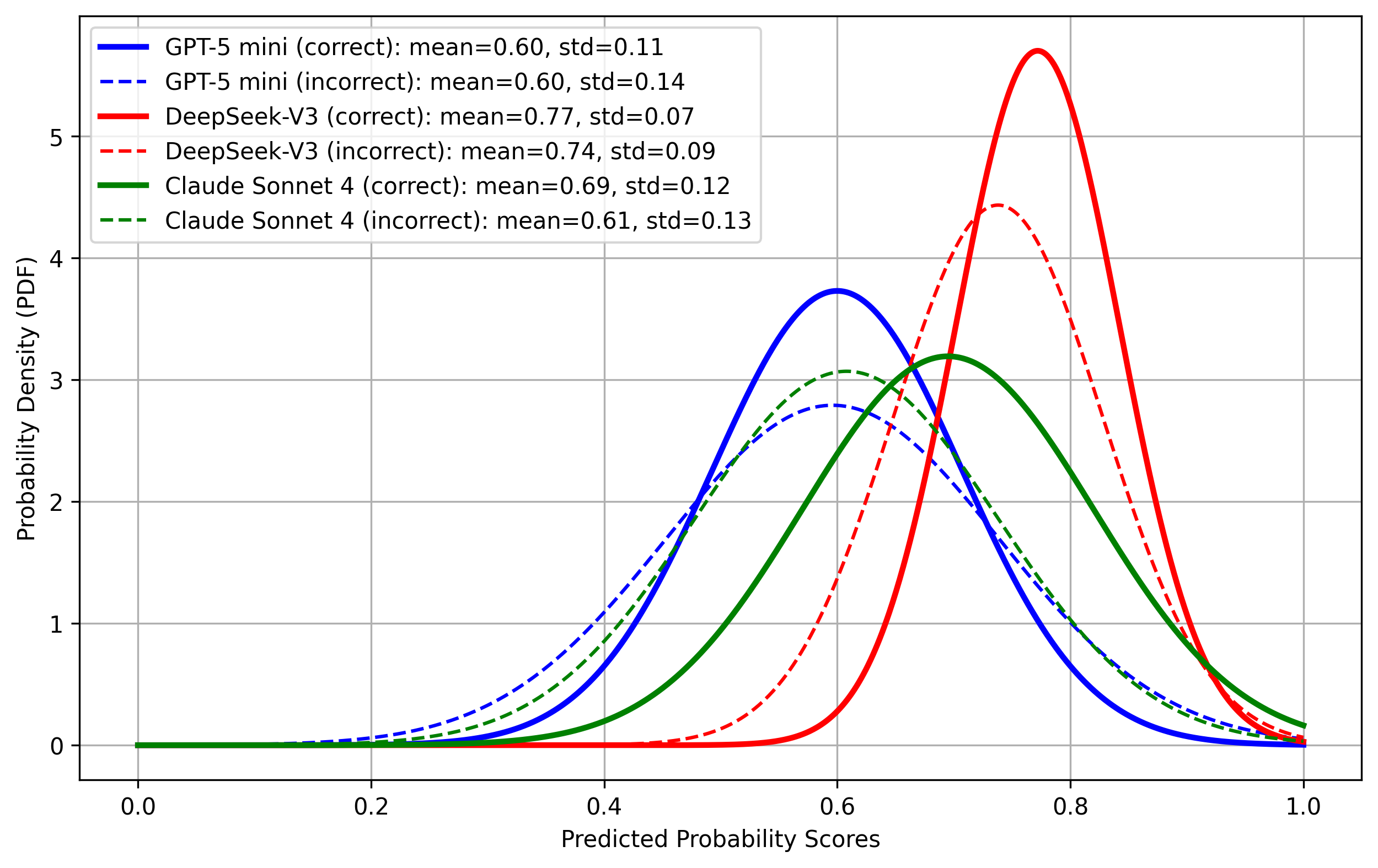}
		\caption{Without Noise.}
		\label{clean_gaussians}
    \end{subfigure}
    \hfill
    \begin{subfigure}[b]{0.45\textwidth}
        \flushleft
		\includegraphics[width=0.99\columnwidth]{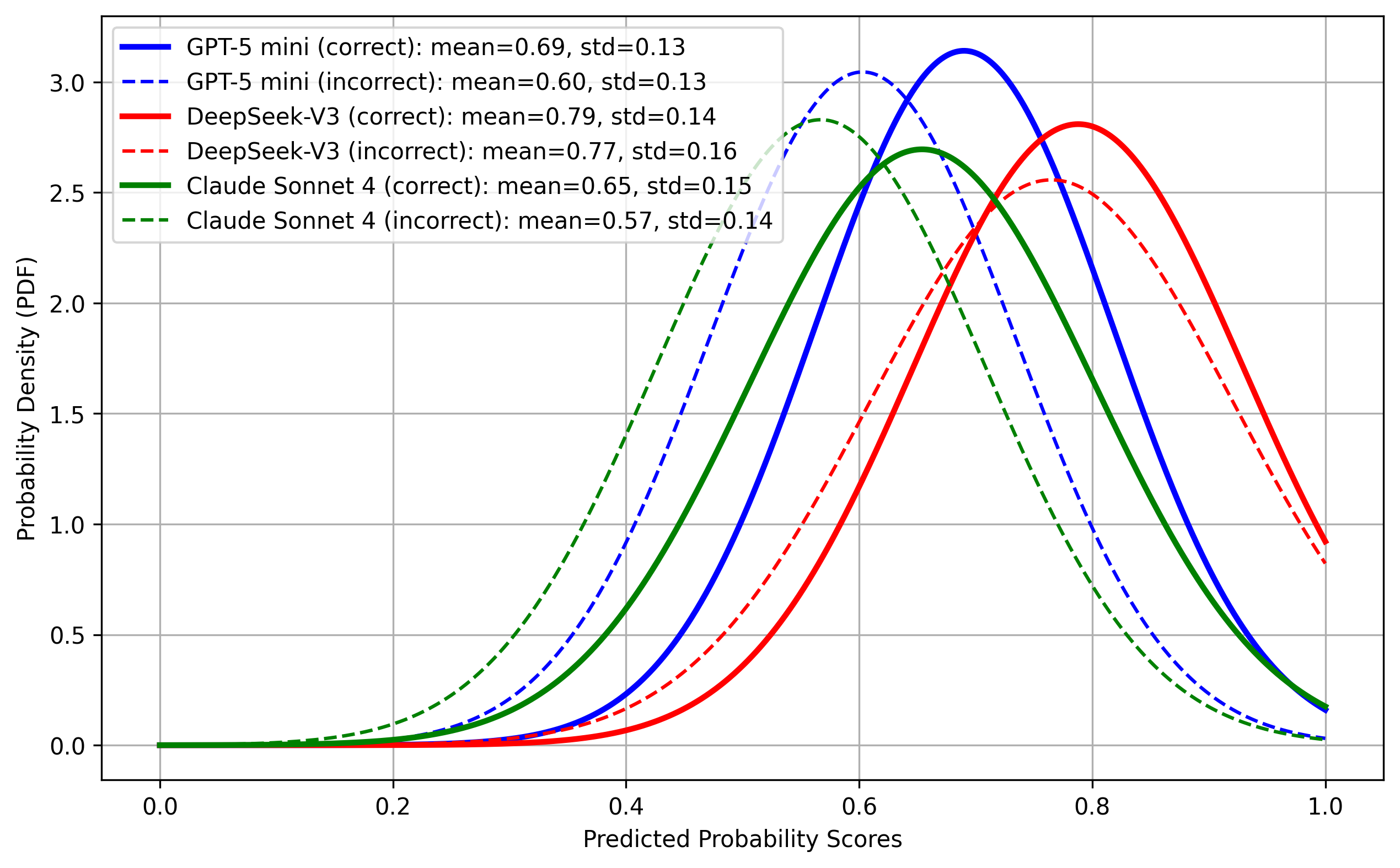}
		\caption{With ASR Noise.}
		\label{noisy_gaussians}
    \end{subfigure}
    \vskip-7pt
    \caption{Gaussian distributions of the assesed LLMs in sentence selection with and without noise; see text for details.} 
    \label{gaussians}
\end{figure*}

\subsection{Benchmark Results of Dialogue Generation}
This benchmark treats dialogue generation as a classification task via sentence selection. Three state-of-the-art LLMs were chosen for this task: GPT-5 mini \cite{ChatGPT}, DeepSeek V3 \cite{DeepSeek}, and Claude Sonnet 4 \cite{Claude}. They were evaluated on the same set of prompts and  template\footnote{Prompt template: \it"Consider the following dialogue context and options. Which option is the most suitable or makes more sense in the dialogue as an AI doctor? Provide your answer with a probability distribution over options, in a single line including option and probability, nothing else." CONTEXT: \$context OPTIONS: \$options}. Whilst variable $\$context$ varied from 5 to 20 dialogue turns in multiples of 4 (i.e., 5, 9, 13, 17 turns), the variable $\$option$ included 20 options--one of them being the correct one, and the remaining options being chosen randomly from all unique doctor sentences in the dataset. This is a challenging goal because the probability of selecting the correct option in each prompt is 1/20=0.05. A subset of doctor-patient dialogues was chosen for this task, which resulted in 1062 prompts without noise and 1104 with noise. 

Whilst manual transcriptions were used for all doctor utterances, noisy dialogues used ASR transcriptions---generated using \href{https://github.com/SYSTRAN/faster-whisper}{Faster-Whisper}---only in the patient utterances.\footnote{The prompts used in this study will be available for reproducibility.}. This is to reflect a real-world application, where the system knows what has been said but it is uncertain on what the patient said (i.e., noisy observations).

Seven performance metrics were used to assess LLM performance: Balanced Accuracy (BA), F1 Score (F1), Area Under the Curve (AUC), Brier Score (BS), KL Divergence (KLD), Binary Cross Entropy (BCE), and Expected Calibration Error (ECE) with 20 bins. The higher the better in the first three, and the lower the better in the last four metrics. 

Table~\ref{llm_metrics} shows results without and with noise,  revealing that {\it Claude Sonnet 4} performs best according to most metrics. 
Results with ASR noise reveal performance degradation in the case of {\it DeepSeek V3}, suggesting that this task is more challenging using ASR transcriptions than manual ones. Interestingly, {\it Claude Sonnet 4} and {\it GPT-5 mini} remained robust against noise, implying that sentence selection may be less sensitive to noise in dialogues than expected. 

Is {\it Claude Sonnet 4} significantly better than the other two LLMs. Yes according to Signed Tests on hard predictions (0s and 1s) at $p<.05$. According to Wilcoxon Signed Rank Tests and T-Tests, all soft predictions of {\it Claude Sonnet 4} were found significantly different from {\it DeepSeek V3} in $p<.05$, but not significantly different from {\it GPT-5 mini} in noisy data. 

\vskip-1pt
\begin{figure}[b!]
    \centering
		\includegraphics[width=0.95\columnwidth]{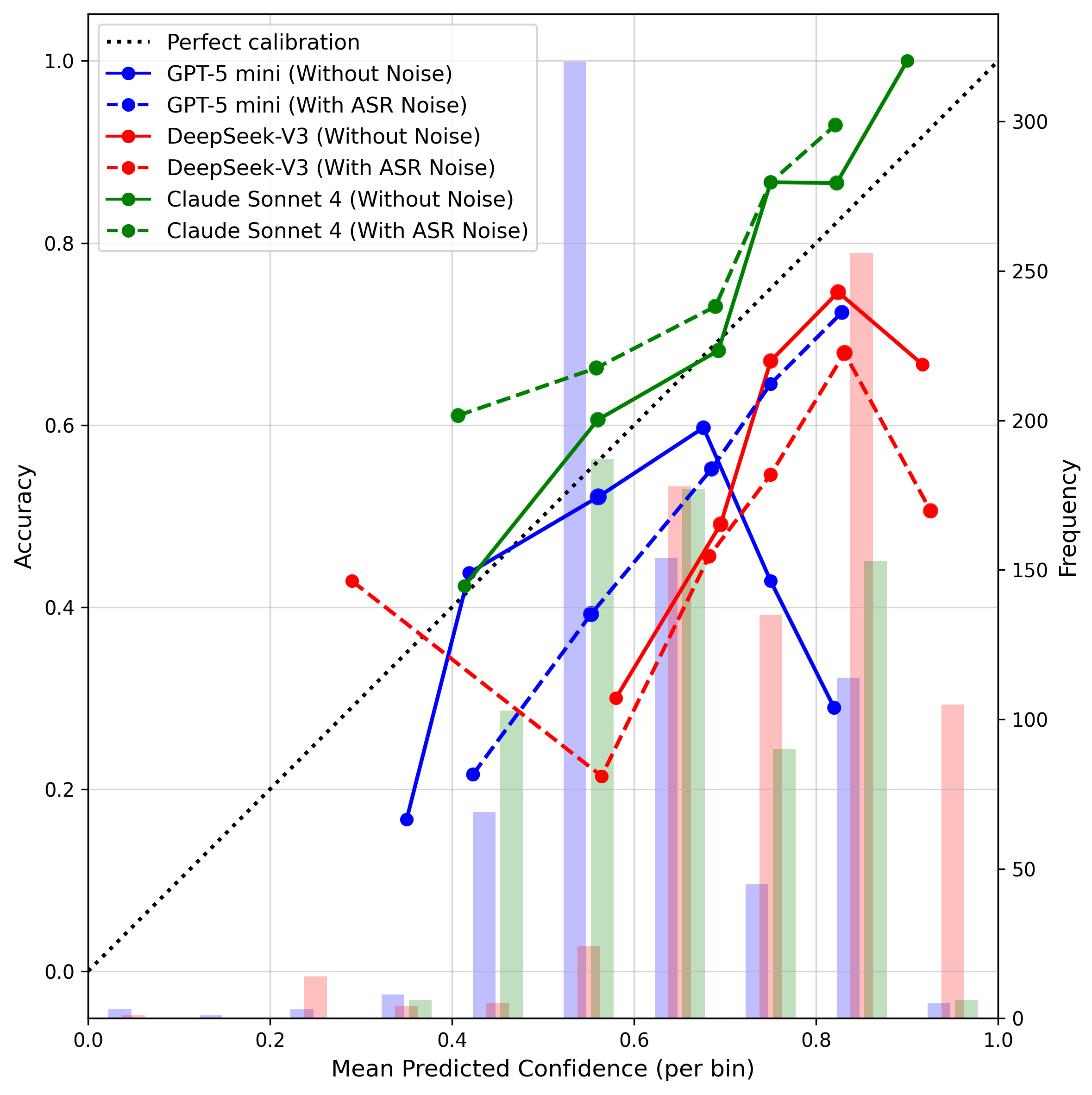}
		\caption{Reliability diagram of the assessed LLMs.}
        \vskip-10pt
		\label{reliability_diagrams}
\end{figure}

Further analysis shows that all LLMs show overconfidence, as can be seen in Figure~\ref{gaussians}. In these two plots, straight lines represent correct predictions and dashed lines represent incorrect predictions. Ideally, a top model should get straight lines on the right-hand side of each plot and dashed lines on the left. However, this is not the case in our study, because our LLMs exhibit nearly equal confidence regardless of whether their selections are correct or incorrect.
\vskip0.2pt
In both plots of Figure~\ref{gaussians} we can observe high levels of overlap in the curves of correct and incorrect predictions per model. According to the Overlap Coefficient (OVL) described in \cite{inman1989overlapping}---where 1 means perfect overlap and 0 no overlap---we obtain the following OVL coefficients for {\it GPT-5 mini}, {\it DeepSeek V3}, and {\it Claude Sonnet 4}, respectively, for dialogues without noise: 0.860, 0.804, 0.734. For dialogues with noise, we obtain the following OVL coefficients for the same order of models: 0.736, 0.927, 0.763. Note that higher OVL values mean more overlap and lower values mean less overlap. Thus, in terms of calibration ability of an LLM, the lower the better. 
Since our computed OVL values exceed 0.7, we observe considerable overlap between the distributions of confidence scores for correct and incorrect predictions. Interestingly, Gaussian plots reveal calibration issues more intuitively than reliability diagrams \cite{MizilCaruana2005icml} as shown in Figure~\ref{reliability_diagrams}. 
Our analysis suggests that the assessed LLMs fail to separate confidence levels between correct and incorrect prediction. This highlights the need for improved model calibration, to achieve predicted probabilities that better align with reality---high for correct predictions and low for incorrect ones.


\section{Conclusions and Future Work}
\label{sec:conclusions}
\vskip-5pt
We present a novel dataset of spoken medical consultations,  collected in realistic environments from robot-patient and doctor-patient dialogues, covering four health conditions: Lewy body dementia, heart failure, shoulder pain, and angina. 
The former included in-person and remote consultations, whereas the latter were carried out face-to-face. About 70\% of spoken dialogues have been annotated with manual transcriptions of doctor and patient utterances, and also with speaker role labels (doctor, robot, patient). It aims to be a useful resource for AI practitioners to develop or evaluate their models using the proposed dataset to eventually create useful applications for real patients. The dialogue benchmark presented in this paper, using sentence selection, suggests that there is room for improvement in all LLMs---especially in the confidence scores assigned when wrong decisions are made. 
Our dataset did not employ data augmentation, which remains to be explored. Further work could provide additional benchmark results for the tasks outlined above in Section~\ref{sec:benchmarks}. 


\section{Acknowledgements}
\vskip-5pt
We would like to acknowledge the \href{https://www.lincoln.ac.uk/medicalschool/}{Lincoln Medical School} for reviewing and approving the ethics applications for collecting this dataset, our 14 BMedSci students for conducting the consultations, our 325 participants who took part as unpaid actor patients, our robot, and our 20 data annotators.

\bibliographystyle{IEEEbib}
\bibliography{strings,refs}

\end{document}